\documentclass[10pt, conference, compsocconf]{IEEEtran}
\usepackage{cite}

\ifCLASSINFOpdf
   \usepackage[pdftex]{graphicx}
\else
\fi
\usepackage[cmex10]{amsmath}
\usepackage[linesnumbered,ruled,vlined]{algorithm2e}

\usepackage{mdwmath}
\usepackage{mdwtab}

\usepackage[caption=false,font=footnotesize]{subfig}
\usepackage{stfloats}

\usepackage{url}

\usepackage{booktabs}
\usepackage{breqn}
\usepackage{footnote}

\makesavenoteenv{tabular}
\makesavenoteenv{table}

\usepackage{balance}
\usepackage{placeins}
\usepackage{bbm}
\usepackage[bookmarks=false]{hyperref}
\usepackage{paralist}
\usepackage{textcomp}
\usepackage{xcolor}
\usepackage{cleveref}

\SetCommentSty{mycommfont}

\SetKwInput{KwInput}{Input}                
\SetKwInput{KwOutput}{Output}              

\def\BibTeX{{\rm B\kern-.05em{\sc i\kern-.025em b}\kern-.08em
 T\kern-.1667em\lower.7ex\hbox{E}\kern-.125emX}}
 
\DeclareMathOperator*{\argmax}{arg\,max}

\newcommand{\norm}[1]{\left\lVert#1\right\rVert}
\newcommand{\fatx}{\mathbf{x}}

\newcommand{\faty}{\mathbf{y}}

\renewcommand{\IEEEbibitemsep}{0pt plus 0.5pt}
\makeatletter
\IEEEtriggercmd{\reset@font\normalfont\fontsize{9pt}{9pt}\selectfont}
\makeatother
\IEEEtriggeratref{1}

\begin{document}
%
\newcommand{\smallmethodname}[1]{{CUDA}}
\newcommand{\fullmethodname}[1]{Contradistinguisher for Unsupervised Domain Adaptation (\smallmethodname{})}
\newcommand{\fullmethodnametitle}[1]{\smallmethodname{}: Contradistinguisher for\\ Unsupervised Domain Adaptation}

\title{CUDA: Contradistinguisher for Unsupervised Domain Adaptation}


\author{\IEEEauthorblockN{Sourabh Balgi\IEEEauthorrefmark{1}
and Ambedkar Dukkipati\IEEEauthorrefmark{2}}
\IEEEauthorblockA{Department of Computer Science and Automation\\
Indian Institute of Science,
Bengaluru, India\\
E-mail: \IEEEauthorrefmark{1}sourabhbalgi@iisc.ac.in,
\IEEEauthorrefmark{2}ambedkar@iisc.ac.in}
}


%


\maketitle

\begin{abstract}
Humans are very sophisticated in learning new information on a completely unknown domain because humans can contradistinguish, i.e., distinguish by contrasting qualities. We learn on a new unknown domain by jointly using unsupervised information directly from unknown domain and supervised information previously acquired knowledge from some other domain. Motivated by this supervised-unsupervised joint learning, we propose a simple model referred as \emph{Contradistinguisher} (CTDR) for unsupervised domain adaptation whose objective is to jointly learn to contradistinguish on unlabeled target domain in a fully unsupervised manner along with prior knowledge acquired by supervised learning on an entirely different domain. Most recent works in domain adaptation rely on an indirect way of first aligning the source and target domain distributions and then learn a classifier on labeled source domain to classify target domain. This approach of indirect way of addressing the real task of unlabeled target domain classification has three main drawbacks. 
\begin{inparaenum}[(i)]
\item The sub-task of obtaining a perfect alignment of the domain in itself might be impossible due to large domain shift (e.g., language domains).
\item The use of multiple classifiers to align the distributions, unnecessarily increases the complexity of the neural networks leading to over-fitting in many cases. 
\item Due to distribution alignment, the domain specific information is lost as the domains get morphed.
\end{inparaenum}
In this work, we propose a simple and direct approach that does not require domain alignment. We jointly learn CTDR on both source and target distribution for unsupervised domain adaptation task using contradistinguish loss for the unlabeled target domain in conjunction with supervised loss for labeled source domain.
Our experiments show that avoiding domain alignment by directly addressing the task of unlabeled target domain classification using CTDR achieves state-of-the-art results on eight visual and four language benchmark domain adaptation datasets.
\end{abstract}


%
\IEEEpeerreviewmaketitle

\section{Introduction}
The recent success of deep neural networks in supervised learning tasks over several areas like computer vision, speech, natural language processing can be attributed to the models that are trained on large amounts of labeled data. However, acquiring large amounts of labeled data in some domains can be very expensive or not possible at all. Additionally, the amount of time required for labeling the data to use existing deep learning techniques can be very high initially for the new domain. This is referred as \emph{cold-start}. On the contrary, cost-effective unlabeled data can be easily obtained in large amounts for most new domains. So, one can aim to transfer the knowledge from a labeled source domain to perform tasks on an unlabeled target domain.

To study this, under the purview of transductive transfer learning, several approaches like domain adaptation, sample selection bias, co-variance shift have been explored in recent times. In this work, we study unsupervised domain adaptation by learning contrastive features in the unlabeled target domain in a fully unsupervised manner utilizing pre-existing informative knowledge from the labeled source domain.
Existing domain adaptation approaches mostly rely on domain alignment, i.e., align both domains so that they are superimposed and indistinguishable. This domain alignment can be achieved in three main ways:
\begin{inparaenum}[(a)] 
\item discrepancy-based methods~\cite{DBLP:conf/iccv/HausserFMC17, 8578490, french2018selfensembling, DBLP:journals/corr/LouizosSLWZ15, 2017arXiv170208811Z},
\item reconstruction-based methods~\cite{10.1007/978-3-319-46493-0_36, Bousmalis:2016:DSN:3157096.3157135}, and
\item adversarial adaptation methods~\cite{pmlr-v37-ganin15, NIPS2016_6544, 8099799, DBLP:conf/cvpr/Sankaranarayanan18a, DBLP:conf/cvpr/LiuYFWCW18, Russo_2018_CVPR, pmlr-v80-hoffman18a, xie2018learning, NIPS2018_7436, DBLP:conf/aaai/ChenCJJ19, shu2018a, hosseini-asl2018augmented}.
\end{inparaenum}

Unlike above methods, our main motivation comes from the human ability to `contradistinguish' and the fundamental idea of statistical learning as described by V. Vapnik~\cite{vapnik1999overview} that indicates any desired problem should be tried to solve in a most possible direct way rather than solving a more general intermediate task. In the context of domain adaptation, the desired problem is classification on the unlabeled target domain and domain alignment followed by most standard methods is the general intermediate. 
This motivates us to propose an approach that does not require domain alignment. 

\noindent Our main contributions in this paper are as follows:
\begin{enumerate}
 \item We propose a simple method that directly addresses the problem of domain adaptation by learning a single classifier, which we refer to as Contradistinguisher (CTDR), jointly in an unsupervised manner over the unlabeled target space and in a supervised manner over the labeled source space. Hence, overcoming the drawbacks of distribution alignment based techniques. 
 \item We formulate a `contradistinguish loss' to directly utilize unlabeled target domain and address the classification task using unsupervised feature learning. A similar approach called DisCoder~\cite{Pandey2017UnsupervisedFL} was used for a much simpler task of semi-supervised feature learning on a single domain with no domain distribution shift.
 \item From our experiments, we show that by jointly training CTDR on the source and target domain distributions, we can achieve above/on-par results over several methods. Surprisingly, this simple method results in improvement over the state-of-the-art for eight challenging benchmark datasets in visual domains (USPS~\cite{lecun1989backpropagation}, MNIST~\cite{lecun1998gradient}, SVHN~\cite{37648}, SYNNUMBERS~\cite{pmlr-v37-ganin15}, CIFAR-10~\cite{krizhevsky2009learning}, STL-10~\cite{coates2011analysis}, SYNSIGNS~\cite{pmlr-v37-ganin15} and GTSRB~\cite{Stallkamp-IJCNN-2011}) and four benchmark language domains (Books, DVDs, Electronics, and Kitchen Appliances) of Amazon customer reviews sentiment analysis dataset~\cite{blitzer2006domain}. 
\end{enumerate}

The rest of the paper is structured as follows. Section \ref{sec:literature review} discusses on related works in domain adaptation. In Section \ref{sec:problem formulation}, we discuss the problem formulation, architecture, loss function definitions, algorithms, and complexity analysis of our proposed method \smallmethodname{}. Section \ref{sec:experiments} deals with the discussion of the experimental setup, results and analysis on vision and language domains. Finally in Section \ref{sec:conclusions and future work}, we conclude by highlighting the key contributions of \smallmethodname{}.

\section{Related Work} \label{sec:literature review}
As mentioned earlier, almost all domain adaptation approaches rely on domain alignment techniques. Here we briefly discuss three main techniques of domain alignment.

\begin{inparaenum}[(a)]
\item \emph{Discrepancy-based methods}:
Associative Domain Adaptation (ADA)~\cite{DBLP:conf/iccv/HausserFMC17} learns statistically domain invariant embeddings using association loss as an alternative to Maximum Mean Discrepancy (MMD)~\cite{Gretton:2009:FCK:2984093.2984169}.
Maximum Classifier Discrepancy (MCD)~\cite{8578490} aligns source and target distributions by maximizing the discrepancy between two separate classifiers.
Self Ensembling (SE)~\cite{french2018selfensembling} uses mean teacher variant~\cite{DBLP:conf/nips/TarvainenV17} of temporal ensembling~\cite{DBLP:conf/iclr/LaineA17} with heavy reliance on data augmentation to minimize the discrepancy between student and teacher network predictions.
Variational Fair Autoencoder (VFAE)~\cite{DBLP:journals/corr/LouizosSLWZ15} uses Variational Autoencoder (VAE)~\cite{DBLP:journals/corr/KingmaW13} with MMD to obtain domain invariant features.
Central Moment Discrepancy (CMD)~\cite{2017arXiv170208811Z} proposes to match higher order moments of source and target domain distributions.

\item \emph{Reconstruction-based methods}:
Deep Reconstruction-Classification Networks (DRCN)~\cite{10.1007/978-3-319-46493-0_36} and Domain Separation Networks (DSN)~\cite{Bousmalis:2016:DSN:3157096.3157135} approaches learn a shared encodings of source and target domains using reconstruction networks.

\item \emph{Adversarial adaptation methods}:
Reverse Gradient (RevGrad/DANN)~\cite{pmlr-v37-ganin15, ganin2016domain} uses domain discriminator to learn domain invariant representations of both the domains.
Coupled Generative Adversarial Network (CoGAN)~\cite{NIPS2016_6544} uses Generative Adversarial Network (GAN)~\cite{Goodfellow:2014:GAN:2969033.2969125} to obtain domain invariant features used for classification.
Adversarial Discriminative Domain Adaptation (ADDA)~\cite{8099799} uses GANs along with weight sharing to learn domain invariant features.
Generate to Adapt (G2A)~\cite{DBLP:conf/cvpr/Sankaranarayanan18a} learns to generate equivalent image in the other domain for a given image, thereby learning common domain invariant embeddings. 
Cross-Domain Representation Disentangler (CDRD)~\cite{DBLP:conf/cvpr/LiuYFWCW18} learns cross-domain disentangled features for domain adaptation.
Symmetric Bi-Directional Adaptive GAN (SBADA-GAN)~\cite{Russo_2018_CVPR} aims to learn symmetric bidirectional mappings among the domains by trying to mimic a target image given a source image. 
Cycle-Consistent Adversarial Domain Adaptation (CyCADA)~\cite{pmlr-v80-hoffman18a} adapts representations at both the pixel-level and
feature-level over the domains.
Moving Semantic Transfer Network (MSTN)~\cite{xie2018learning} proposes moving semantic transfer network that learn semantic representations for the unlabeled target samples by aligning labeled source centroids and pseudo-labeled target centroids.
Conditional Domain Adversarial Network (CDAN)~\cite{NIPS2018_7436} conditions the adversarial adaptation models on discriminative information conveyed in the classifier predictions.
Joint Discriminative Domain Adaptation (JDDA)~\cite{DBLP:conf/aaai/ChenCJJ19} proposes joint domain alignment along with discriminative feature learning.
Decision-boundary Iterative Refinement Training with a Teacher (DIRT-T)~\cite{shu2018a} and Augmented Cyclic Adversarial Learning (ACAL)~\cite{hosseini-asl2018augmented} learn by using a domain discriminator along with data augmentation for domain adaptation.
\end{inparaenum}

Apart from these standard ways, a slight deviant method explored is \emph{Tri-Training}. Tri-Training algorithms use three classifiers trained on the labeled source domain and refine them for unlabeled target domain. To be precise, in each round of tri-training, a target sample is pseudo-labeled if the other two classifiers agree on the labeling, under certain conditions such as confidence thresholding. Asymmetric Tri-Training (ATT)~\cite{pmlr-v70-saito17a} uses three classifiers to bootstrap high confidence target domain samples by confidence thresholding. This way of bootstrapping works only if the source classifier has very high accuracy. In case of of low source classifier accuracy, target samples are never obtained to bootstrap, resulting in a bad model. 
Multi-Task Tri-training (MT-Tri)~\cite{DBLP:conf/acl/PlankR18} explores the tri-training technique on the language domain adaptation tasks.

All the domain adaptation approaches mentioned earlier have a common unifying theme: they attempt to morph the target and source distributions so as to make them indistinguishable. Once the two distributions are perfectly aligned, they use a classifier trained on labeled source domain to classify the unlabeled target domain. Hence, the performance of the classifier on the target domain depends crucially on the domain alignment. As a result, the actual task of target domain classification is solved indirectly using domain alignment rather than using the unlabeled target data in an unsupervised manner which is a more logical and direct way.

In this paper, we propose a completely different approach: instead of focusing on aligning the source and target distributions, we learn a single classifier referred as \emph{Contradistinguisher} (CTDR), jointly on both the domain distributions using contradistinguish loss for the unlabeled target data and supervised loss for the labeled source data. 

\section{Proposed Method: \smallmethodname{}} \label{sec:problem formulation}
A domain $\mathcal{D}_d$ is specified by its input feature space $\mathcal{X}_d$, the label space $\mathcal{Y}_d$ and the joint probability distribution $p(\fatx_d, \faty_d)$, where $\fatx_d{ \in }\mathcal{X}_d$ and $\faty_d{ \in }\mathcal{Y}_d$. Let $\left \lvert \mathcal{Y}_d \right \rvert{ = }K$ be the number of class labels such that $\faty_d{ \in }\{0, \ldots, K{ - }1\}$ for any instance $\fatx_d$. In particular, Domain adaptation consists of two domains $\mathcal{D}_{s}$ and $\mathcal{D}_{t}$ that are referred as the source and target domains respectively. A common assumption in domain adaptation is that the input feature space as well as the label space remains unchanged across the source and the target domain, i.e., $\mathcal{X}_s{ = }\mathcal{X}_t{ = }\mathcal{X}_d$ and $\mathcal{Y}_s{ = }\mathcal{Y}_t{ = }\mathcal{Y}_d$. Hence, the only difference between the source and target domain is input-label space distributions, i.e., $p(\fatx_s,\faty_s){ \neq }p(\fatx_t,\faty_t)$. This is referred as \emph{domain shift} in the standard literature of domain adaptation.

In particular, in an unsupervised domain adaptation, the training data consists of labeled source domain instances ${\{(\fatx^{i}_s, \faty^{i}_s)\}}^{n_s}_{i=1}$ and unlabeled target domain instances ${\{\fatx^{j}_t\}}^{n_t}_{j=1}$. Given a labeled data in the source domain, it is straightforward to learn a classifier by maximizing the conditional probability $p(\faty_s|\fatx_s)$ over the labeled samples. However, the task at hand is to learn a classifier on the unlabeled target domain by transferring the knowledge from the labeled source domain.

\subsection{Overview} \label{sec:overview}
\begin{figure}[t!]
\centering
\includegraphics[width=0.9\linewidth,height=0.9\textheight,keepaspectratio=true]{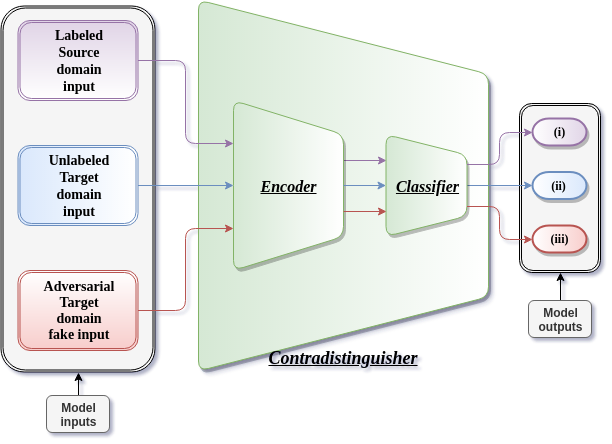}
\caption{Architecture of the proposed method \smallmethodname{} with \emph{Contradistinguisher} (\emph{Encoder} and \emph{Classifier}). Three optimization objectives with their respective inputs involved in training of \smallmethodname{}: (i) Source supervised~\eqref{eq:ce}, (ii) Target unsupervised~\eqref{eq:unsup}, and Adversarial regularization~\eqref{eq:adv reg loss}.}
\label{fig:training method}
\end{figure}
Figure~\ref{fig:training method} indicates the model architecture of our proposed method~\smallmethodname{}, i.e., \emph{Contradistinguisher} (CTDR) and the respective losses involved in \smallmethodname{} training. 

The objective of CTDR is to find a clustering scheme using the most contrastive features on unlabeled target in such a way that it also satisfies the target domain prior, i.e., \emph{prior enforcing}. We achieve this by jointly training labeled source samples in a supervised manner and unlabeled target samples in an unsupervised end-to-end manner by using a contradistinguish loss same as~\cite{Pandey2017UnsupervisedFL}. This fine-tunes the classifier learnt from source domain to the target domain. The main important feature of our approach is 
the contradistinguish loss~\eqref{eq:unsup} which is discussed in detail in Section~\ref{sec:unsupervised target classification}.

Note that the objective of the CTDR is not same as a classifier, i.e., distinguishing is not same as classifying.
Suppose there are two contrastive entities $e_1{ \in }C_1$ and $e_2{ \in }C_2$, where $C_1, C_2$ are two classes. The aim of a classifier is to classify $e_1{ \in }C_1$ and $e_2{ \in }C_2$, where to train a classifier one requires labeled data. On the contrary, the job of a CTDR is to just identify $e_1{ \neq }e_2$, i.e., CTDR can classify $e_1{ \in }C_1$ (or $C_2$) and $e_2{ \in }C_2$ (or $C_1$) indifferently. To train CTDR, we do not need any class information but only need unlabeled entities $e_1$ and $e_2$. Using unlabeled target data, CTDR is able to distinguish the samples in an unsupervised way. However, since the final task is classification, one would require a selective incorporation of the pre-existing informative knowledge required for the task of classification. This knowledge is obtained by jointly training, thus classifying $e_1{ \in }C_1$ and $e_2{ \in }C_2$.

In the subsequent Sections \ref{sec:supervised source classification}--\ref{sec:complexity analysis}, we formulate the optimization objectives and also perform complexity analysis.

\subsection{Supervised Source Classification} \label{sec:supervised source classification}
For the labeled source domain instances ${\{(\fatx^{i}_s, \faty^{i}_s)\}}^{n_s}_{i=1}$, we define the conditional-likelihood of observing $\faty_s$ given $\fatx_s$ as, $p_{\theta}(\faty_{s}|\textbf{x}_{s})$,
where $\theta$ denotes the parameters of CTDR.

We estimate $\theta$ by maximizing the conditional log-likelihood of observing the labels given the labeled source domain samples. The source domain supervised objective to maximize
\setlength{\arraycolsep}{0.0em}
\begin{eqnarray}
\mathcal{L}_{s}(\theta)&{}={}&\sum_{i=1}^{n_s}\log(p_{\theta}(\faty^{i}_{s}|\textbf{x}^{i}_{s}))\enspace.
\label{eq:sup}
\end{eqnarray}
\setlength{\arraycolsep}{5pt}
Alternatively, one can minimize the cross-entropy loss
\setlength{\arraycolsep}{0.0em}
\begin{eqnarray}
\mathcal{L}_{ce}(\theta)&{}={}&-\sum_{i=1}^{n_s}\sum_{k=0}^{K-1}{\mathbbm{1}[\faty^{i}_s{ = }k]}\log(\hat{\faty}^{ik}_{s})\enspace,
\label{eq:ce}
\end{eqnarray}
where 
$\hat{\faty}^{ik}_{s}$ is the softmax output of CTDR that represents the probability of class $k$ for the given sample $\fatx^{i}_s$.

\subsection{Unsupervised Target Classification} \label{sec:unsupervised target classification}
For the unlabeled target domain instances ${\{\fatx^{j}_t\}}^{n_t}_{j=1}$, as the corresponding labels are unknown, a naive way of predicting the target labels is to directly use the classifier trained only with supervised loss~\eqref{eq:ce}. Though this gives some good results, it fails to achieve high accuracies due to two reasons:
\begin{inparaenum}[(i)] 
\item $p_{\theta}$ is defined over $\fatx_s$ and not $\fatx_t$.
\item $p_{\theta}$ is not a valid probability distribution because $\sum_{\ell=1}^{n_t}p_{\theta}(\faty_t|\textbf{x}_t^{\ell}) \neq 1$.
\end{inparaenum}
Enforcing these two conditions, we model a non-trivial joint distribution $\hat{q}_\theta(\textbf{x}_t,\faty_t)$ parameterized by $\theta$ over target domain as,
\setlength{\arraycolsep}{0.0em}
\begin{eqnarray}
\hat{q}_{\theta}(\textbf{x}_t,\faty_t)&{}={}&\frac{p_{\theta}(\faty_t|\textbf{x}_t)}{\sum_{\ell=1}^{n_t}p_{\theta}(\faty_t|\textbf{x}_t^{\ell})}\enspace.
\label{eq:mod unsup1}
\end{eqnarray}
\setlength{\arraycolsep}{5pt}
However~\eqref{eq:mod unsup1} is not exactly a joint distribution yet because $\sum_{\ell=1}^{n_t}\hat{q}_{\theta}(\textbf{x}^{\ell}_{t}, \faty_{t}){ \neq }p(\faty_t)$, i.e., marginalizing over all ${\{\fatx^{j}_t\}}^{n_t}_{j=1}$ should yield the target prior distribution $p(\faty_t)$. 
We modify~\eqref{eq:mod unsup1} so as to include the marginalization condition. We refer to this as \emph{target domain prior enforcing}.
\setlength{\arraycolsep}{0.0em}
\begin{eqnarray}
q_{\theta}(\textbf{x}_t,\faty_t)&{}={}&\frac{p_{\theta}(\faty_t|\textbf{x}_t){*}p(\faty_t)}{\sum_{\ell=1}^{n_t}p_{\theta}(\faty_t|\textbf{x}_t^{\ell})}\enspace.
\label{eq:mod unsup}
\end{eqnarray}
\setlength{\arraycolsep}{5pt}
Note that $q_\theta(\fatx_t, \faty_t)$ defines a non-trivial approximate of joint distribution over the target domain as a function of $p_\theta$ learnt over source domain. The resultant unsupervised maximization objective for the target domain is given by maximizing the log-probability of the joint distribution $q_\theta(\fatx_t, \faty_t)$ which is
\setlength{\arraycolsep}{0.0em}
\begin{eqnarray}
\mathcal{L}_{t}(\theta, {\{\faty^{j}_t\}}^{n_t}_{j=1})&{}={}&\sum_{j=1}^{n_t}\log(q_{\theta}(\textbf{x}^{j}_{t}, \faty^{j}_{t}))\enspace.
\label{eq:unsup}
\end{eqnarray}
\setlength{\arraycolsep}{5pt}
Next, we discuss how the objective~\eqref{eq:unsup} is solved and the reason why~\eqref{eq:unsup} is referred as contradistinguish loss.
Since the target labels ${\{\faty^{j}_t\}}^{n_t}_{j=1}$ are unknown, one needs to maximize~\eqref{eq:unsup} over the parameters $\theta$ as well as the unknown target labels $\faty_t$. As there are two parameters for maximization, we follow a two step approach to maximize~\eqref{eq:unsup}. The two optimization steps are as follows.

\begin{inparaenum}[(i)] 
\item \emph{Pseudo-label selection}: We maximize \eqref{eq:unsup} only with respect to the label $\faty_t$ for every $\fatx_t$ by fixing $\theta$ as
\setlength{\arraycolsep}{0.0em}
\begin{eqnarray}
\hat{\faty}^{j}_t&{}={}&\argmax_{\faty^{j} \in \mathcal{Y}_t} \frac{p_{\theta}(\faty^{j}|\textbf{x}^{j}_t){ * }p(\faty_t)}{\sum_{\ell=1}^{n_t}p_{\theta}(\faty^{\ell}| \textbf{x}_t^{\ell})}\enspace.
\label{eq:pseudo label selection}
\end{eqnarray}
\setlength{\arraycolsep}{5pt}
\noindent
Pseudo-labeling approach under semi-supervised representation learning setting has been well studied in~\cite{pseudo-label} and shown equivalent to \emph{entropy regularization}~\cite{grandvalet2005semi}. We derive the motivation from~\cite{Pandey2017UnsupervisedFL} that also use pseudo-labeling in the context of semi-supervised representation learning. However, our method addresses a more complex problem of domain adaptation in the presence of domain shift.

\item \emph{Maximization}: By fixing the pseudo-labels ${\{\hat{\faty}^{j}_t\}}^{n_t}_{j=1}$ from~\eqref{eq:pseudo label selection}, we train CTDR to maximize \eqref{eq:unsup} with respect to the parameter $\theta$.
\setlength{\arraycolsep}{0.0em}
\begin{eqnarray}
\mathcal{L}_t(\theta)&{}={}&\sum_{j=1}^{n_t}\log(p_{\theta}(\hat{\faty}^{j}_t|\textbf{x}^{j}_t)) + \sum_{j=1}^{n_t}\log(p(\faty_t))\nonumber\\
&&{-}\:\sum_{j=1}^{n_t}\log(\sum_{\ell=1}^{n_t}p_{\theta}(\hat{\faty}^{\ell}_t| \textbf{x}_t^{\ell}))\enspace.
\label{eq:unsup full}
\end{eqnarray}
\setlength{\arraycolsep}{5pt}
\noindent
The first term, i.e., log-probability for a given $\fatx^{j}_t$ forces CTDR to choose features to classify $\fatx^{j}_t$ to $\hat{\faty}^{j}_t$. The second term is a constant, hence it has no effect in optimization with respect to $\theta$. The third term is the negative of log-probability for all the samples $\fatx_t$ in the entire domain. Maximization of this term forces CTDR to choose features to not classify all the other $\fatx^{\ell{ \neq }j}_t$ to selected pseudo-label $\hat{\faty}^{j}_t$ except the given sample $\fatx^{j}_t$. This forces CTDR to extract the most unique features of a given sample $\fatx^{j}_t$ against all the other samples $\fatx^{\ell{ \neq }j}_t$, i.e., most unique contrastive feature of the selected sample $\fatx^{j}_t$ over all the other samples $\fatx^{\ell{ \neq }j}_t$ to distinguish a given sample from all others.

The first and third term together in~\eqref{eq:unsup full} enforce that CTDR learns the most contradistinguishing features among the samples $\fatx_t{ \in }\mathcal{X}_t$, thus performing unlabeled target domain classification in a fully unsupervised way. Because of this contradistinguishing feature learning, we refer the unsupervised target domain objective \eqref{eq:unsup} as contradistinguish loss.
\end{inparaenum}

Ideally, one would like to compute the third term in~\eqref{eq:unsup full} using the complete target training data for each input sample. Since it is expensive to compute the third term over the entire $\fatx_t$ for each individual sample during training, one evaluates the third term in~\eqref{eq:unsup full} over a mini-batch. In our experiments, we have observed that mini-batch strategy does not cause any problem during training as far as it includes at least one sample from each class which is guaranteed for a reasonably large mini-batch size of $128$. For numerical stability, we use $\log\sum\exp$ trick to optimize third term in~\eqref{eq:unsup full}.

\subsection{Adversarial Regularization} \label{sec:adversarial regularization}
In order to prevent CTDR from over-fitting to the chosen pseudo labels during the training, we use adversarial regularization. In particular, we train CTDR to be confused about set of fake $-ve$ samples ${\{\hat{\fatx}^{j}_t\}}^{n_f}_{j=1}$ by maximizing the conditional log-probability over the given fake sample such that the sample belongs to all $K(\left \lvert \mathcal{Y}_d \right \rvert)$ classes simultaneously. The objective of the adversarial regularization is to multi-label the fake sample (e.g., noisy image that looks like a cat and a dog) equally to all $K$ classes as labeling to any unique class introduces more noise in pseudo labels. This strategy is similar to entropy regularization~\cite{grandvalet2005semi} in the sense that instead of minimizing the entropy for the real target samples, we maximize the conditional log-probability over the fake $-ve$ samples. Therefore, we add the following maximization objective to the total CTDR objective as a regularizer. 
\setlength{\arraycolsep}{0.0em}
\begin{eqnarray}
\mathcal{L}_{adv}(\theta)&{}={}&\sum_{j=1}^{n_f}\log(p_{\theta}(\hat{\faty}_t^{j}|\hat{\fatx}^{j}_{t}))\enspace,
\label{eq:adv reg}
\end{eqnarray}
for all $\hat{\faty}^{j}_t{ \in }\mathcal{Y}_t$.
As maximization of \eqref{eq:adv reg} is analogous to minimize the binary cross-entropy loss \eqref{eq:adv reg loss} of a multi-class multi-label classification task, in our practical implementation, we minimize \eqref{eq:adv reg loss} for assigning labels to all the classes for every samples.
\setlength{\arraycolsep}{0.0em}
\begin{eqnarray}
\mathcal{L}_{bce}(\theta)&{}={}&-\sum_{j=1}^{n_f}\sum_{k=0}^{K-1}\log(\hat{\faty}^{jk}_{t})\enspace,
\label{eq:adv reg loss}
\end{eqnarray}
where 
$\hat{\faty}^{jk}_{t}$ is the softmax output of CTDR which represents the probability of class $k$ for the given sample $\hat{\fatx}^{j}_t$.

The fake $-ve$ samples $\hat{\fatx}_t$ can be directly sampled from, say a Gaussian distribution in the input feature space $\mathcal{X}_t$ with the mean and standard deviation of the samples $\fatx_t{ \in }\mathcal{X}_t$. For the language domain, fake samples are generated randomly as mentioned above. 
In case of image datasets, as the feature space is high dimensional, the fake images $\hat{\fatx}_t$ are generated using a generator network $G_{\phi}$ with parameter $\phi$ that takes Gaussian noise vector $\eta_t$ as input to produce a fake sample $\hat{\fatx}_t$, i.e., $\hat{\fatx}_t = G_{\phi}(\eta_t)$. Generator $G_{\phi}$ is trained by minimizing kernel MMD loss~\cite{DBLP:conf/nips/LiCCYP17}, i.e., a modified version of MMD loss between the encoder output $\rho(\hat{\fatx}_t)$ and $\rho(\fatx_t)$ of $n_f$ fake images $\hat{\fatx}_t$ and $n_t$ real target domain images $\fatx_t$ respectively. 
\setlength{\arraycolsep}{0.0em}
\begin{eqnarray}
\mathcal{L}_{gen}(\phi)&{}={}&\frac{1}{n_f^{2}}\sum_{i=1}^{n_f}\sum_{j=1}^{n_f} k(\rho(\hat{\fatx}_t^{i}),\rho(\hat{\fatx}_t^{j}))\nonumber\\
&&{+}\:\frac{1}{n_t^{2}}\sum_{i=1}^{n_t}\sum_{j=1}^{n_t} k(\rho(\fatx_t^{i}),\rho(\fatx_t^{j}))\nonumber\\
&&{-}\:\frac{2}{n_t*n_f}\sum_{i=1}^{n_f}\sum_{j=1}^{n_t} k(\rho(\hat{\fatx}_t^{i}),\rho(\fatx_t^{j})),
\label{eq:kMMD loss gen}
\end{eqnarray}
\setlength{\arraycolsep}{5pt}
\noindent where $k(x,x') = e^{-\gamma\norm{x-x'}^{2}}$ is the Gaussian kernel.
 
Note that the objective of the generator is not to generate realistic image but to generate fake noisy images with mixed image attributes from the target domain. This reduces the effort of training powerful generators which is the focus in adversarial based domain adaptation approaches~\cite{DBLP:conf/cvpr/Sankaranarayanan18a, DBLP:conf/cvpr/LiuYFWCW18, Russo_2018_CVPR, pmlr-v80-hoffman18a, xie2018learning} used for domain alignment.


\subsection{Algorithms and Complexity Analysis} \label{sec:complexity analysis}
Algorithm~\ref{alg: method training} and~\ref{alg: method inference} list steps involved in \smallmethodname{} training and inference respectively. 
\begin{algorithm}[t!]
\DontPrintSemicolon
  
  \KwInput{$b{ = }batch\_size$, $epochs{ = }max\_epoch$, $n_{batch}{ = } number\ of\ batches$}
  \KwOutput{$\theta$ \tcp*{parameter of CTDR}}
  \KwData{${\{(\fatx^{i}_s, \faty^{i}_s)\}}^{n_s}_{i=1}$, ${\{\fatx^{j}_t\}}^{n_t}_{j=1}$}
    \If{target domain prior $p(\faty_t)$ is known}
    {
        use $p(\faty_t)$ for the contradistinguish loss~\eqref{eq:unsup}  
    }
    \Else
    {
    	compute $p(\faty_t)$ assuming $p(\faty_t) = p(\faty_s)$  
	}
    \For{$epoch = 1$ to $epochs$}
    {
        \For{$batch = 1$ to $n_{batch}$}
        {
            sample a mini-batch ${\{(\fatx^{i}_s, \faty^{i}_s)\}}^{b}_{i=1}$, ${\{\fatx^{j}_t\}}^{b}_{j=1}$
            
            compute $\mathcal{L}_{s}(\theta)$~\eqref{eq:sup}\label{alg_stt: source supervised loss} using ${\{(\fatx^{i}_s, \faty^{i}_s)\}}^{b}_{i=1}$
            
            compute ${\{\hat{\faty}^{j}_t\}}^{b}_{j=1}$~\eqref{eq:pseudo label selection} using ${\{\fatx^{j}_t\}}^{b}_{j=1}$  
            
            compute $\mathcal{L}_{t}(\theta)$~\eqref{eq:unsup full} fixing ${\{\hat{\faty}^{j}_t\}}^{b}_{j=1}$ \label{alg_stt: target unsupervised loss}
            
            \If{adversarial regularization is enabled}
            {
                \If{Generator $G_\phi$ is used}
                {
                get fake samples ${\{\hat{\fatx}^{j}_t\}}^{b}_{j=1}$ from Gaussian noise vectors ${\{\eta^{j}_t\}}^{b}_{j=1}$ using $G_\phi$, compute $\mathcal{L}_{gen}(\phi)$\eqref{eq:kMMD loss gen}\label{alg_stt: generator loss}  
                }
                \Else
                {
                get fake samples ${\{\hat{\fatx}^{j}_t\}}^{b}_{j=1}$ by random sampling in the input feature space $\mathcal{X}_t$
                }
                compute $\mathcal{L}_{adv}(\theta)$~\eqref{eq:adv reg loss} using ${\{\hat{\fatx}^{j}_t\}}^{b}_{j=1}$ \label{alg_stt: adv loss} 
            }
            combine losses in steps~\ref{alg_stt: source supervised loss},~\ref{alg_stt: target unsupervised loss},~\ref{alg_stt: generator loss}, and~\ref{alg_stt: adv loss} to compute gradients using backward-pass
            
            update $\theta$ using gradient descent 
        }
    }
\caption{\smallmethodname{} Training}
\label{alg: method training}
\end{algorithm}
\begin{algorithm}[t!]
\DontPrintSemicolon
  
  \KwInput{${\{\fatx^{i}_{test}\}}^{n_{test}}_{i=1}$  \tcp*{input test samples}}
  \KwOutput{${\{\hat{\faty}^{i}_{test}\}}^{n_{test}}_{i=1}$ \tcp*{predicted labels}}

    \For{$i = 1$ to $n_{test}$}
    {
        
        predict label as $\hat{\faty}^{i}_{test}{}={}\argmax_{\faty \in \mathcal{Y}_t} p_{\theta}(\faty|\textbf{x}^{i}_{test})$\label{alg_stt: predict labels}
    }
\caption{\smallmethodname{} Inference}
\label{alg: method inference}
\end{algorithm}
Here we briefly discuss time complexity of Algorithm~\ref{alg: method training} and~\ref{alg: method inference}. We also compare model complexity of \smallmethodname{} against domain alignment approaches.

\begin{inparaenum}[(a)]
\item \emph{Time complexity}:
We consider a batch of $b$ instances for forward and backward propagation during training. For source supervised loss~\eqref{eq:ce}, the time complexity is $O(b K T_c)$, where $T_c$ is the time complexity involved in obtaining the classifier output which mainly depends on the model complexity which will be discussed next. 
For target unsupervised loss~\eqref{eq:unsup}, the time complexity is $O(b K T_c)$ for pseudo-label selection and $O(b K T_c + b^2 K T_c)$ for first and third terms in maximization step, i.e., $O(b^2 K T_c)$ effectively for the target unsupervised loss~\eqref{eq:unsup}. 
The adversarial regularization loss~\eqref{eq:adv reg loss} complexity corresponds to $O(b K T_c)$.
Time complexity for generator training is $O(b^2 D_e T_e)$, where $D_e$ is dimension of the encoder output and $T_e$ is the time complexity for the encoder output from neural network which also depends on the model complexity discussed next.
~As $T_c$ dominates $T_e$, total training time complexity can be further simplified to \textbf{$O(b^2 K T_c)$}.
During inference phase, the time complexity is \textbf{$O(n_{test} T_c)$}, where $n_{test}$ is the number of inference samples. 

\item \emph{Model complexity}:
As discussed above, $T_c$ mainly depends on the model complexity involving many factors such as input feature dimension, number of neural network layers, type of normalization, type of activation functions etc. CTDR is a simple network with a single encoder and classifier unlike MCD-DA that uses a single encoder with two classifier. This makes MCD-DA time complexity $2 T_c$ instead of just $T_c$. Similarly, SE uses 2 copies of network of encoder and classifier one for student and other for teacher network. This makes SE time complexity $2 T_c$ instead of $T_c$. In general, as domain alignment approaches use additional circuitry either in-terms of multiple classifiers or GANs, the model complexity increases at least by a factor of 2. This increased model complexity requires more data augmentation to prevent under-fitting leading to further increases in time complexity at the expense of only a slight improvement, if any, compared to \smallmethodname{} as indicated by our state-of-the-art results without any data augmentation in both visual and language domain adaptation tasks. 
We observed empirically, most of the computational complexity is for the forward and backward propagation to obtain the classifier softmax output and the gradients, i.e., $T_c$. Hence the use of GPUs to accelerate $T_c$.
We believe the trade-off achieved by the simplicity of \smallmethodname{}, as evident from our results, is very desirable compared to most domain alignment approaches that use data augmentation and complex neural networks for a slight improvement, if any.
\end{inparaenum}

\section{Experiments} \label{sec:experiments}
\begin{figure}[t!]
 \begin{center}
 \includegraphics[width=0.9\linewidth,height=0.9\textheight,keepaspectratio=true]{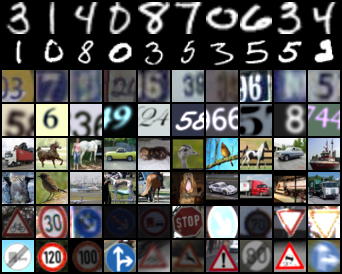}
 \caption{Illustrations of random samples from
 (a) $\mathcal{US}$, (b) $\mathcal{MN}$, (c) $\mathcal{SV}$, (d) $\mathcal{SN}$, (e) $\mathcal{C}9$, (f) $\mathcal{S}9$, (g) $\mathcal{GT}$ and (h) $\mathcal{SS}$. 
(from top to bottom row.)}
 \label{fig:datasets}
 \end{center}
\end{figure}
\begin{table*}[!t]
    \begin{minipage}{0.6\linewidth}
    \caption{Details of image datasets.}
    \begin{center}
    \label{tbl:visual datasets}
    \setlength\tabcolsep{5pt}
    \begin{tabular}{lrrrccc}
        \toprule
    Dataset & \# Train & \# Test & \# Classes & Target & Resolution & Channels\\
    \midrule
    USPS ($\mathcal{US}$) & 7,291 & 2,007 & 10 & Digits & 16 $\times$ 16 & Mono\\ 
    MNIST ($\mathcal{MN}$) & 60,000 & 10,000 & 10 & Digits & 28 $\times$ 28 & Mono\\ 
    SVHN ($\mathcal{SV}$) & 73,257 & 26,032 & 10 & Digits & 32 $\times$ 32 & RGB\\ 
    SYNNUMBERS ($\mathcal{SN}$) & 479,400 & 9,553 & 10 & Digits & 32 $\times$ 32 & RGB\\ 
    \midrule
    CIFAR-9 ($\mathcal{C}9$) & 45,000 & 9,000 & 9 & Object ID & 32 $\times$ 32 & RGB\\ 
    STL-9 ($\mathcal{S}9$) & 4,500 & 7,200 & 9 & Object ID & 96 $\times$ 96 & RGB\\ 
    \midrule
    SYNSIGNS ($\mathcal{SS}$) & 100,000 & - & 43 & Traffic Signs & 40 $\times$ 40 & RGB\\ 
    GTSRB ($\mathcal{GT}$) & 39,209 & 12,630 & 43 & Traffic Signs & varies & RGB\\ 
    \bottomrule
    \end{tabular}
    \end{center}
    \end{minipage}
    \hfill
    \begin{minipage}{.3\linewidth}
    \caption{Details of language dataset (Amazon customer reviews 
    for sentiment analysis).}
    \begin{center}
    \label{tbl:lang datasets}
    \setlength\tabcolsep{5pt}
 \begin{tabular}{lrr}
        \toprule
    Domain & \# Train & \# Test\\
    \midrule
    Books ($\mathcal{B}$) & 2,000 & 4,465\\
    DVDs ($\mathcal{D}$) & 2,000 & 3,586\\
    Electronics ($\mathcal{E}$) & 2,000 & 5,681\\
    Kitchen Appliances ($\mathcal{K}$) & 2,000 & 5,945\\
    \bottomrule
    \end{tabular}
    \end{center}
    \end{minipage} 
\end{table*}
\begin{figure*}[ht!]
\begin{center}
\subfloat[Before training]{
\includegraphics[width=0.45\linewidth,height=0.325\linewidth,keepaspectratio=true]{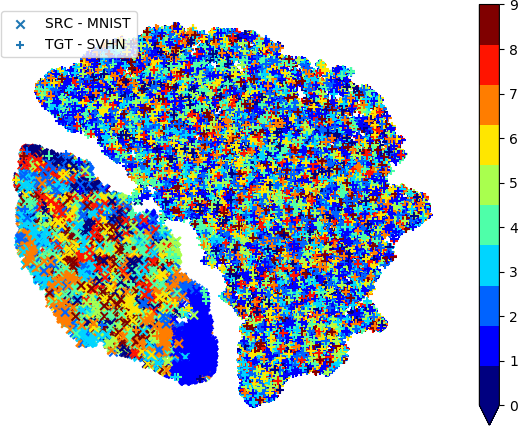}
\label{sfig:tsne svhn mnist a}
}  
\hfill
\subfloat[after 1 $epoch$ training]{
\includegraphics[width=0.45\linewidth,height=0.45\textheight,keepaspectratio=true]{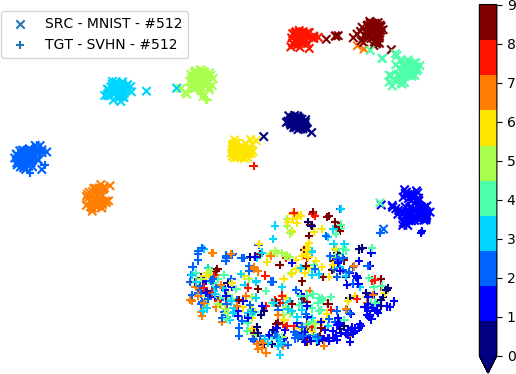}
\label{sfig:tsne svhn mnist b}
}
\hfill
\subfloat[after 6 $epochs$ training]{
\includegraphics[width=0.45\linewidth,height=0.45\textheight,keepaspectratio=true]{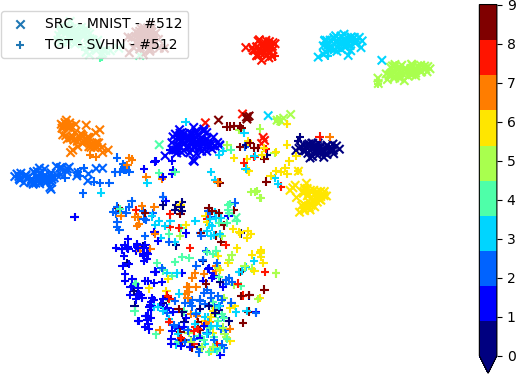}
\label{sfig:tsne svhn mnist c}
}
\hfill
\subfloat[after full training]{
\includegraphics[width=0.45\linewidth,height=0.45\textheight,keepaspectratio=true]{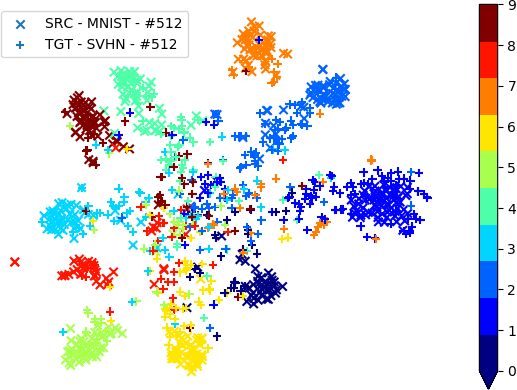}
\label{sfig:tsne svhn mnist d}
}
\caption{t-SNE~\cite{vandermaaten2008visualizing} plots for embeddings from the output of CTDR before applying softmax corresponding to the test samples from $\mathcal{MN}{ \rightarrow }\mathcal{SV}$ visual task trained with \smallmethodname{}. We consider this task as this is the most difficult among all the visual experiments due contrasting domains with high domain shift.
(a) Initial plot of all the test samples before training indicating domain shift as there are two separate clusters for each domain. 
(b) Plot of subset from test samples after $epoch{=}1$. 
(c) Plot of subset from test samples after $epoch{=}6$. 
(d) Plot of subset from test samples after full \smallmethodname{} training. 
}
\label{fig:tsne svhn mnist}
\end{center}
\end{figure*}
\begin{figure*}[ht!]
\begin{center}
\subfloat[USPS$ \rightarrow $MNIST]{
\includegraphics[width=0.215\linewidth,height=0.215\textheight,keepaspectratio=true]{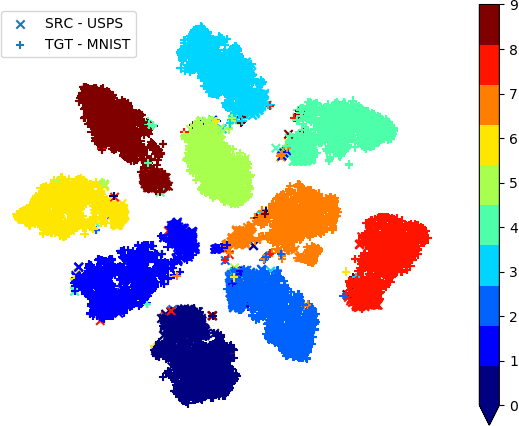}
}
\hfill
\subfloat[MNIST$ \rightarrow $USPS]{
\includegraphics[width=0.215\linewidth,height=0.215\textheight,keepaspectratio=true]{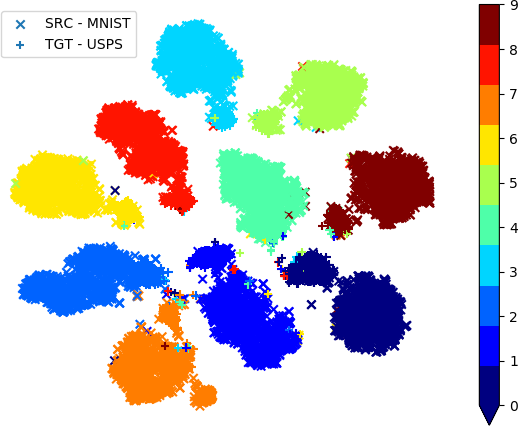}}
\hfill
\subfloat[SVHN$ \rightarrow $MNIST]{
\includegraphics[width=0.215\linewidth,height=0.215\textheight,keepaspectratio=true]{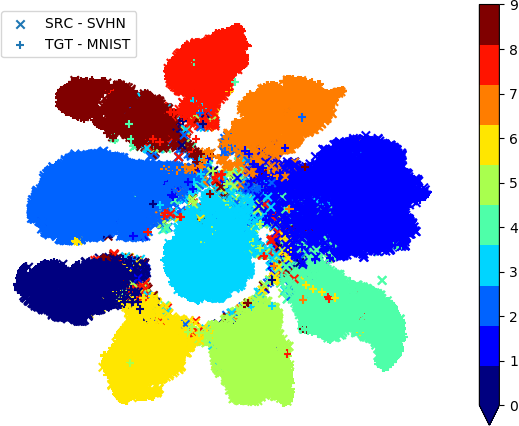}}
\hfill
\subfloat[MNIST$ \rightarrow $SVHN]{
\includegraphics[width=0.215\linewidth,height=0.215\textheight,keepaspectratio=true]{mnist_svhn_clip_epoch20_noacc_cut.png}}
\\
\subfloat[CIFAR-9$ \rightarrow $STL-9]{
\includegraphics[width=0.215\linewidth,height=0.215\textheight,keepaspectratio=true]{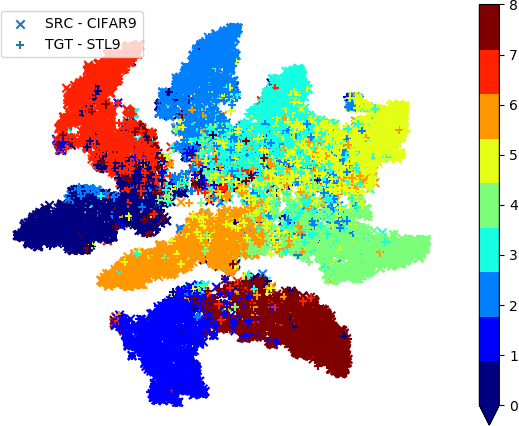}}
\hfill
\subfloat[STL-9$ \rightarrow $CIFAR-9]{
\includegraphics[width=0.215\linewidth,height=0.215\textheight,keepaspectratio=true]{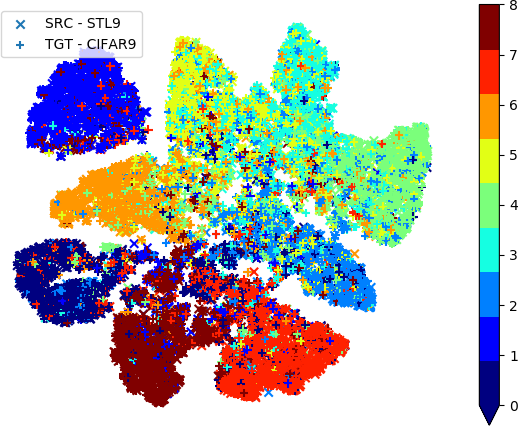}}
\hfill
\subfloat[SYNNUMBERS$ \rightarrow $SVHN]{
\includegraphics[width=0.215\linewidth,height=0.215\textheight,keepaspectratio=true]{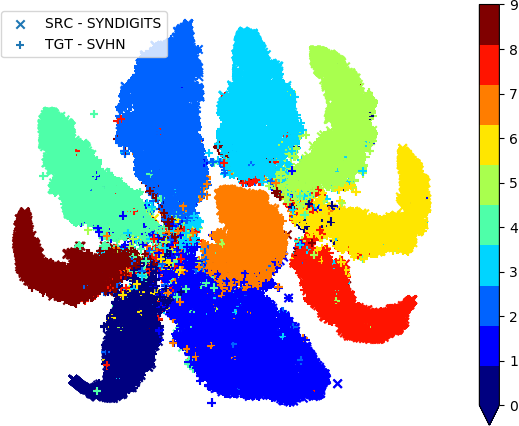}}
\hfill
\subfloat[SYNSIGNS$ \rightarrow $GTSRB]{
\includegraphics[width=0.215\linewidth,height=0.215\textheight,keepaspectratio=true]{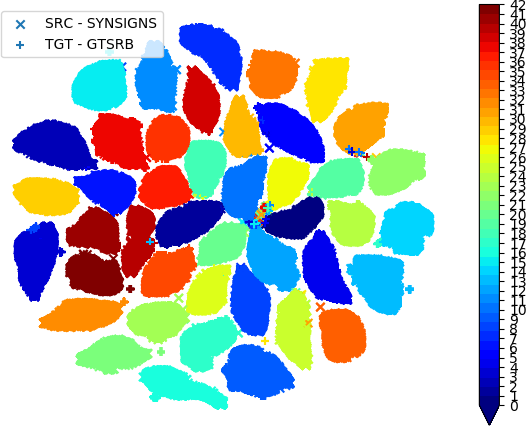}}
\caption{t-SNE~\cite{vandermaaten2008visualizing} plots for embeddings from the output of CTDR before applying softmax corresponding to the test samples in visual experiments. 
}
\label{fig:tsne visual all}
\end{center}
\end{figure*}
\begin{table*}[ht!]
\centering
\caption{Target domain test accuracy (\%) on image datasets. 
\smallmethodname{} corresponds to our best results obtained with best hyper-parameter settings. 
$ts{ = }\mathcal{BL}_1$: target supervised, $ss{ = }\mathcal{BL}_2$: source supervised, $tu$: target unsupervised, $su$: source unsupervised, $ta$: adversarial regularization and $sa$: source adversarial regularization represents different training configurations. 
We exclude~\cite{french2018selfensembling, shu2018a, hosseini-asl2018augmented} from comparison as they use heavy data augmentation.
}
\label{tbl:image results}
 \setlength\tabcolsep{7pt}
\begin{tabular}{lcccccccc}
\toprule
\textbf{Method} & $\mathcal{US}{ \rightarrow }\mathcal{MN}$ & $\mathcal{MN}{ \rightarrow }\mathcal{US}$ & $\mathcal{SV}{ \rightarrow }\mathcal{MN}$ & $\mathcal{MN}{ \rightarrow }\mathcal{SV}$ & $\mathcal{C}9{ \rightarrow }\mathcal{S}9$ & $\mathcal{S}9{ \rightarrow }\mathcal{C}9$ & $\mathcal{SN}{ \rightarrow }\mathcal{SV}$ & $\mathcal{SS}{ \rightarrow }\mathcal{GT}$ \\
\midrule
ADA~\cite{DBLP:conf/iccv/HausserFMC17} & - & - & 97.16 & - & - & - & 91.86 & 97.66 \\
MCD-DA~\cite{8578490} & 94.10 & 94.20 & 96.20 & - & - & - & - & 94.40\\ 
DRCN~\cite{10.1007/978-3-319-46493-0_36} & 73.67 & 91.80 & 81.97 & 40.05 & 66.37 & 58.65 & - & - \\ 
DSN~\cite{Bousmalis:2016:DSN:3157096.3157135} & - & - & 82.70 & - & - & - & 91.20 & 93.10 \\ 
RevGrad~\cite{pmlr-v37-ganin15} & 74.01 & 91.11 & 73.91 & 35.67 & 66.12 & 56.91 & 91.09 & 88.65 \\ 
CoGAN~\cite{NIPS2016_6544} & 89.10 & 91.20 & - & - & - & - & - & - \\
ADDA~\cite{8099799} & 90.10 & 89.40 & 76.00 & - & - & - & - & - \\
G2A~\cite{DBLP:conf/cvpr/Sankaranarayanan18a} & 90.80 & 92.50 & 84.70 & 36.40 & - & - & - & - \\ 
CDRD~\cite{DBLP:conf/cvpr/LiuYFWCW18} & 94.35 & 95.05 & - & - & - & - & - & - \\ 
SBADA-GAN~\cite{Russo_2018_CVPR} & 95.00 & 97.60 & 76.10 & 61.10 & - & - & - & 96.70 \\ 
CyCADA~\cite{pmlr-v80-hoffman18a} & 96.50 & 95.60 & 90.40 & - & - & - & - & - \\
MSTN~\cite{xie2018learning} & - & 92.90 & 91.70 & - & - & - & - & - \\ 
CDAN~\cite{NIPS2018_7436} & 97.10 & 96.50 & 90.50 & - & - & - & - & - \\ 
JDDA~\cite{DBLP:conf/aaai/ChenCJJ19} & 96.70 & - & 94.20 & - & - & - & - & - \\ 
ATT~\cite{pmlr-v70-saito17a} & - & - & 86.20 & 52.80 & - & - & 93.10 & 96.20 \\ 
\midrule
\textbf{\smallmethodname{}} (Ours) & \textbf{99.20} & \textbf{97.86} & \textbf{99.07} & \textbf{71.30} & \textbf{77.22} & \textbf{65.93} & \textbf{94.30} & \textbf{99.40}\\ 
\midrule
\midrule
$ts{ = }\mathcal{BL}_1$ (Ours) & 99.64 & 97.98 & 99.64 & 96.02 & 73.78 & 91.46 & 96.85 & 98.23 \\ 
$ss{ = }\mathcal{BL}_2$ (Ours) & 81.18 & 82.00 & 77.54 & 24.86 & 77.64 & 62.10 & 91.45 & 95.13 \\ 
\midrule
$ss{ + }tu$ (Ours) & 98.83 & 97.71 & 98.81 & 50.83 & \textbf{77.22} & \textbf{62.50} & \textbf{93.65} & 98.15 \\ 
$ss{ + }tu{ + }su$ (Ours) & 98.77 & \textbf{97.86} & 98.62 & 54.38 & 76.93 & 61.09 & 93.52 & 97.86 \\ 
$ss{ + }tu{ + }su{ + }ta$ (Ours) & \textbf{99.20} & 97.31 & \textbf{98.85} & 54.32 & 76.18 & 59.37 & 93.59 & \textbf{99.40} \\ 
$ss{ + }tu{ + }su{ + }sa$ (Ours) & 89.97 & 93.87 & 97.15 & 41.71 & 75.00 & 56.99 & 90.79 & 99.35 \\ 
$ss{ + }tu{ + }su{ + }sa{ + }ta$ (Ours) & 98.75 & 96.26 & 95.73 & \textbf{55.25} & 70.93 & 61.37 & 92.97 & 99.11 \\ 
\midrule
\midrule
SE~\cite{french2018selfensembling} & 99.54 & 98.26 & 99.26 & 97.00 & 80.09 & 74.24 & 97.11 & 99.37 \\ 
DIRT-T~\cite{shu2018a} & - & - & 99.40 & 54.50 & - & 73.30 & 96.20 & 99.60 \\ 
ACAL~\cite{hosseini-asl2018augmented} & 97.16 & 98.31 & 96.51 & 60.85 & - & - & 97.98 & - \\ 
\bottomrule
\end{tabular}
\end{table*}
\begin{figure*}[ht!]
\begin{center}
\subfloat[$\mathcal{B} \rightarrow \mathcal{K}$]{
\includegraphics[width=0.215\linewidth,height=0.215\textheight,keepaspectratio=true]{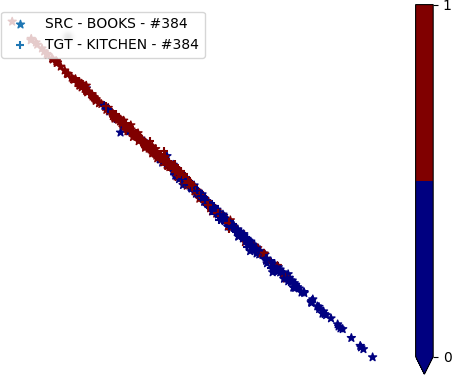}}
\hfill
\subfloat[$\mathcal{D} \rightarrow \mathcal{K}$]{
\includegraphics[width=0.215\linewidth,height=0.215\textheight,keepaspectratio=true]{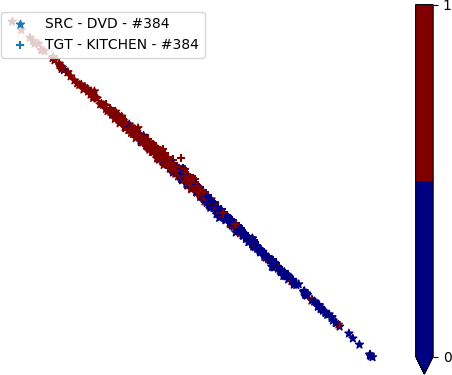}}
\hfill
\subfloat[$\mathcal{E} \rightarrow \mathcal{K}$]{
\includegraphics[width=0.215\linewidth,height=0.215\textheight,keepaspectratio=true]{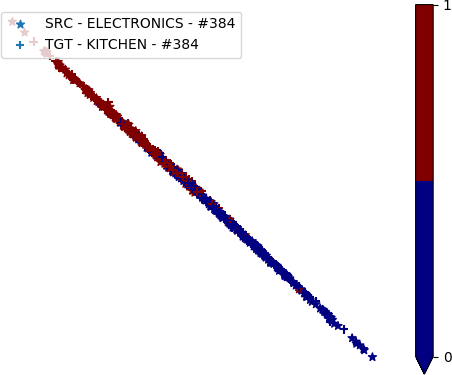}}
\hfill
\subfloat[$\mathcal{K} \rightarrow \mathcal{E}$]{
\includegraphics[width=0.215\linewidth,height=0.215\textheight,keepaspectratio=true]{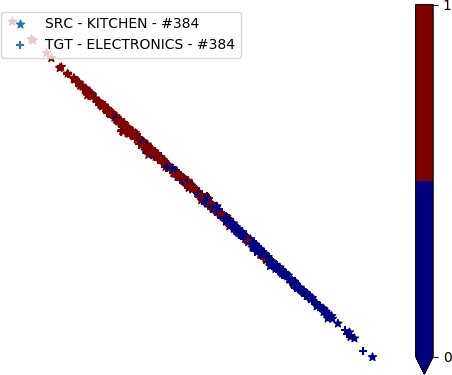}}
\caption{t-SNE~\cite{vandermaaten2008visualizing} plots for embeddings from the output of CTDR corresponding to the test samples in language experiments . 
(Note: For the sake of the brevity, we do not add the plots for all the language tasks as language tasks plots are almost similar and not as informative as visual tasks.)
}
\label{fig:tsne all lang}
\end{center}
\end{figure*}
\begin{table*}[ht!]
\centering
 \caption{Target domain test accuracy (\%) on Amazon customer reviews dataset for Sentiment Analysis. 
 \smallmethodname{} corresponds to our best results obtained with best hyper-parameter settings. 
 $ts{ = }\mathcal{BL}_1$: target supervised, $ss{ = }\mathcal{BL}_2$: source supervised, $tu$: target unsupervised, $su$: source unsupervised and $ta$: adversarial regularization represents different training configurations. 
 }
 \label{tbl:text results}
 \setlength\tabcolsep{5pt}
 \begin{tabular}{lccccccccccccc}
 \toprule
 \textbf{Method} & $\mathcal{B}{ \rightarrow }\mathcal{D}$ & $\mathcal{B}{ \rightarrow }\mathcal{E}$ & $\mathcal{B}{ \rightarrow }\mathcal{K}$ & $\mathcal{D}{ \rightarrow }\mathcal{B}$ & $\mathcal{D}{ \rightarrow }\mathcal{E}$ & $\mathcal{D}{ \rightarrow }\mathcal{K}$ & $\mathcal{E}{ \rightarrow }\mathcal{B}$ & $\mathcal{E}{ \rightarrow }\mathcal{D}$ & $\mathcal{E}{ \rightarrow }\mathcal{K}$ & $\mathcal{K}{ \rightarrow }\mathcal{B}$ & $\mathcal{K}{ \rightarrow }\mathcal{D}$ & $\mathcal{K}{ \rightarrow }\mathcal{E}$ & $Mean$\\
 \midrule
 VFAE~\cite{DBLP:journals/corr/LouizosSLWZ15} & 79.90 & 79.20 & 81.60 & 75.50 & 78.60 & 82.20 & 72.70 & 76.50 & 85.00 & 72.00 & 73.30 & 83.80 & 78.35 \\
 CMD~\cite{2017arXiv170208811Z} & 80.50 & 78.70 & 81.30 & 79.50 & 79.70 & 83.00 & 74.40 & 76.30 & 86.00 & \textbf{75.60} & 77.50 & 85.40 & 79.82 \\
DANN~\cite{ganin2016domain} & 78.40 & 73.30 & 77.90 & 72.30 & 75.40 & 78.30 & 71.30 & 73.80 & 85.40 & 70.90 & 74.00 & 84.30 & 76.27 \\ 
  ATT~\cite{pmlr-v70-saito17a} & 80.70 & 79.80 & 82.50 & 73.20 & 77.00 & 82.50 & 73.20 & 72.90 & 86.90 & 72.50 & 74.90 & 84.60 & 78.39 \\
 MT-Tri~\cite{DBLP:conf/acl/PlankR18} & 78.14 & 81.45 & 82.14 & 74.86 & 81.45 & 82.14 & 74.86 & \textbf{78.14} & 82.14 & 74.86 & \textbf{78.14} & 81.45 & 79.14 \\
 \midrule
 \textbf{\smallmethodname{}} (Ours) & \textbf{82.77} & \textbf{83.07} & \textbf{85.58} & \textbf{80.02} & \textbf{82.06} & \textbf{85.70} & \textbf{75.88} & 76.05 & \textbf{87.30} & 73.08 & 73.06 & \textbf{86.66} & \textbf{80.93} \\
 \midrule
 \midrule
 
 $ts{ = }\mathcal{BL}_1$ (Ours) & 83.83 & 87.19 & 89.05 & 84.08 & 87.19 & 89.05 & 84.08 & 83.83 & 89.05 & 84.08 & 83.83 & 87.19 & 86.03 \\
 $ss{ = }\mathcal{BL}_2$ (Ours) & 81.07 & 75.11 & 77.53 & 77.67 & 75.99 & 79.78 & 73.12 & 74.48 & 86.19 & 72.59 & 76.24 & 85.92 & 77.97 \\
 \midrule
 $ss{ + }tu$ (Ours) & 81.99 & 81.45 & 84.36 & 77.18 & 81.48 & 84.37 & 67.26 & 67.71 & 87.30 & 70.68 & 71.97 & 84.79 & 78.37 \\
  $ss{ + }tu{ + }su$ (Ours) & 82.63 & 81.73 & 83.75 & 75.88 & 77.45 & 80.96 & 69.70 & 70.69 & 87.37 & 72.99 & 67.76 & 84.51 & 77.91 \\
  $ss{ + }tu{ + }su{ + }ta$ (Ours) & \textbf{82.77} & \textbf{83.07} & \textbf{85.58} & \textbf{80.02} & \textbf{82.06} & \textbf{85.70} & \textbf{75.88} & \textbf{76.05} & 87.30 & \textbf{73.08} & \textbf{73.06} & \textbf{86.66} & \textbf{80.93} \\
  $ss{ + }tu{ + }su{ + }sa{ + }ta$ (Ours) & 80.37 & 80.20 & 84.58 & 78.45 & 81.36 & 85.03 & 75.05 & 75.01 & \textbf{87.47} & 72.63 & 71.97 & 86.31 & 79.86 \\
 \bottomrule
\end{tabular}
\end{table*}

\subsection{Experimental Setup} \label{sec:setup}
\subsubsection{Visual Domain Adaptation} \label{sec: visual exp setup}
We consider eight benchmark visual datasets
with 3 different nature of images for our visual domain experiments. 
\begin{inparaenum}[(a)]
\item Digits:
USPS ($\mathcal{US}$)~\cite{lecun1989backpropagation} and 
MNIST ($\mathcal{MN}$)~\cite{lecun1998gradient} 
are a pair of gray-scale digits datasets. 
SVHN ($\mathcal{SV}$)~\cite{37648} and  
SYNNUMBERS ($\mathcal{SN}$)~\cite{pmlr-v37-ganin15} are another pair of RGB digits datasets. 
\item Objects:
CIFAR ($\mathcal{C}9$)~\cite{krizhevsky2009learning} and 
STL ($\mathcal{S}9$)~\cite{coates2011analysis} 
are a dataset pair of objects/animals RGB images by considering only the 9 overlapping classes from the original datasets.
\item Traffic Signs:
SYNSIGNS ($\mathcal{SS}$)~\cite{pmlr-v37-ganin15} and  
GTSRB ($\mathcal{GT}$)~\cite{Stallkamp-IJCNN-2011}
~are a dataset pair with traffic signs.
\end{inparaenum}
Table~\ref{tbl:visual datasets} provides visual dataset details and Figure~\ref{fig:datasets} indicates some random samples from all eight datasets. 

On these datasets, we consider eight main domain adaptation tasks studied in~\cite{pmlr-v37-ganin15, french2018selfensembling}. These eight visual tasks and the data processing considered are as follows,
\begin{inparaenum}[(i)]
\item $\mathcal{US}{ \leftrightarrow }\mathcal{MN}$: $\mathcal{US}$ images are up-scaled using bi-linear interpolation from 16$\times$16$\times$1 to 28$\times$28$\times$1 to match the size of $\mathcal{MN}$,
\item $\mathcal{SV}{ \leftrightarrow }\mathcal{MN}$: $\mathcal{MN}$ images are up-scaled using bi-linear interpolation to 32$\times$32$\times$1. The RGB channels of $\mathcal{SV}$ are converted to Mono image resulting in 32$\times$32$\times$1 size. Several other combinations were tried and this was chosen since the results are the best,
\item $\mathcal{SN}{ \rightarrow }\mathcal{SV}$: No pre-processing required as these domains have same image size,
\item $\mathcal{C}9{ \leftrightarrow }\mathcal{S}9$: Only the 9 overlapping classes from datasets as the label space should be same for both the domain. $\mathcal{S}9$ images are down-scaled from 96$\times$96$\times$3 to 32$\times$32$\times$3 to match the size of $\mathcal{C}9$.
\item $\mathcal{SS}{ \rightarrow }\mathcal{GT}$: Crop the images to 40$\times$40$\times$3 based on the region of interest in the images in both datasets.
\end{inparaenum}

Note that we do not perform any image data augmentation in our experiments unlike~\cite{french2018selfensembling}.
Our aim in this paper is to demonstrate that the proposed method performs above/on-par without data augmentation as data augmentation is expensive and not always possible as seen in language tasks. 

\subsubsection{Language Domain Adaptation} \label{sec: language exp setup}
We consider four benchmark language domains \begin{inparaenum}[(i)]
\item Books ($\mathcal{B}$),
\item DVDs ($\mathcal{D}$),
\item Electronics ($\mathcal{E}$), and
\item Kitchen Appliances ($\mathcal{K}$)
\end{inparaenum}
from Amazon customer reviews~\cite{blitzer2006domain} dataset.
The dataset includes product reviews in four different domains for sentiment analysis as indicated in Table~\ref{tbl:lang datasets}. 

On these domains, we consider all twelve tasks studied in~\cite{ganin2016domain, DBLP:journals/corr/LouizosSLWZ15, 2017arXiv170208811Z, pmlr-v70-saito17a, DBLP:conf/acl/PlankR18}.
We use the same neural networks and text pre-processing used in~\cite{Chen:2012:MDA:3042573.3042781, ganin2016domain, DBLP:conf/acl/PlankR18} to get 5000 dimensional feature vector. We assign binary label `0' for the products rated from $\leq3$ stars and `1' for $\geq4$ star ratings. 

We select the best existing neural networks without major modifications to hyper-parameters so as to demonstrate the effectiveness of \smallmethodname{}. All the experiments are done using PyTorch~\cite{paszke2017automatic} with mini-batch size of 64 per GPU distributed over four GPUs, Adam optimizer with an initial learning rate $0.001$ and decay rate of $0.6$ every 30 epochs. 

\subsection{Experimental Results} \label{sec:results}
We use the same metric used for evaluation as in~\cite{pmlr-v37-ganin15, NIPS2016_6544, 10.1007/978-3-319-46493-0_36, pmlr-v70-saito17a, DBLP:conf/iccv/HausserFMC17, 8099799, DBLP:conf/cvpr/Sankaranarayanan18a, DBLP:conf/cvpr/LiuYFWCW18, 8578490, Russo_2018_CVPR, pmlr-v80-hoffman18a, xie2018learning, NIPS2018_7436, DBLP:conf/aaai/ChenCJJ19, french2018selfensembling, shu2018a, hosseini-asl2018augmented, ganin2016domain, DBLP:journals/corr/LouizosSLWZ15, 2017arXiv170208811Z, DBLP:conf/acl/PlankR18, Bousmalis:2016:DSN:3157096.3157135}, i.e., the accuracy on target domain test set. 
Table~\ref{tbl:image results} indicates the target domain test accuracy across all the eight main domain adaptation tasks compared with several state-of-the-art domain alignment methods~\cite{pmlr-v37-ganin15, NIPS2016_6544, 10.1007/978-3-319-46493-0_36, pmlr-v70-saito17a, DBLP:conf/iccv/HausserFMC17, 8099799, DBLP:conf/cvpr/Sankaranarayanan18a, DBLP:conf/cvpr/LiuYFWCW18, 8578490, Russo_2018_CVPR, pmlr-v80-hoffman18a, xie2018learning, NIPS2018_7436, DBLP:conf/aaai/ChenCJJ19, french2018selfensembling, shu2018a, hosseini-asl2018augmented, Bousmalis:2016:DSN:3157096.3157135}. 
Table~\ref{tbl:text results} indicates the target domain test accuracy across all the twelve domain adaptation tasks compared with different state-of-the-art methods~\cite{ganin2016domain, DBLP:journals/corr/LouizosSLWZ15, 2017arXiv170208811Z, pmlr-v70-saito17a, DBLP:conf/acl/PlankR18}.

Apart from the standard domain alignment methods used for comparison, we report two baselines $\mathcal{BL}_1$ and $\mathcal{BL}_2$ of our own, reported in Tables~\ref{tbl:image results} and~\ref{tbl:text results}, by fixing the CTDR neural network architecture and varying only the training losses used to demonstrate the effectiveness of \smallmethodname{}. $\mathcal{BL}_1$ indicates training CTDR using only the target domain in a fully supervised way. $\mathcal{BL}_2$ indicates training CTDR using only the source domain in a fully supervised way. 
$\mathcal{BL}_1$ and $\mathcal{BL}_2$ respectively indicates the maximum and minimum target domain test accuracy that can be attained with chosen CTDR neural network.

Comparing \smallmethodname{} with $\mathcal{BL}_2$ in Tables~\ref{tbl:image results} and~\ref{tbl:text results}, we can see huge improvements in the target domain test accuracies due to the use of contradistinguish loss~\eqref{eq:unsup} demonstrating the effectiveness of CTDR.

As our method is mainly dependent on the contradistinguish loss~\eqref{eq:unsup}, experimenting with better neural networks along with our contradistinguish loss~\eqref{eq:unsup}, we observed better results in both visual and language domain adaptation task over the neural networks used in~\cite{8099799, 8578490} on visual experiments and MAN~\cite{DBLP:conf/naacl/ChenC18} on language experiments.


\subsection{Analysis of Experimental Results} \label{sec:analysis}

\subsubsection{Visual Domain Adaptation} \label{sec: visual exp analysis}

In tasks $\mathcal{C}9{ \rightarrow }\mathcal{S}9$ and $\mathcal{SS}{ \rightarrow }\mathcal{GT}$, $\mathcal{BL}_1{ < }$\smallmethodname{} in Table~\ref{tbl:image results}. $\mathcal{BL}_1$ is poor because $n_t{ \ll }n_s$ causing under-fitting during only target domain supervised loss training. The improved results of \smallmethodname{} indicates that CTDR is able contradistinguish on the target domain along with the transfer of informative knowledge required for the classification from a larger source domain. This indicates that CTDR is indeed successful in contradistinguishing on a relatively small set of unlabeled target domain using larger source domain information. 
Other interesting observation is in the task $\mathcal{C}9{ \rightarrow }\mathcal{S}9$, where $\mathcal{BL}_2$ is slightly better than \smallmethodname{}. This is due to slight over-fitting on the target domain training examples which are actually non-informative for classification leading to a small decrease in the target domain test accuracy. $\mathcal{BL}_2{ > }\mathcal{BL}_1$ indicates source domain has more information than target domain due to large source and small target training sets.
~Figure~\ref{fig:tsne svhn mnist}(a-d) shows t-SNE plots for $\mathcal{MN}{ \rightarrow }\mathcal{SV}$ as the training progresses using \smallmethodname{}. 
We indicate these plots as this is the most difficult among all the visual experiments due contrasting domains.
~Figure~\ref{fig:tsne visual all}(a-h) shows t-SNE plots on the test sample outputs of CTDR for all eight visual experiments and they show clear class-wise clustering on both source and target domains indicating the efficacy of \smallmethodname{}.

\subsubsection{Language Domain Adaptation} \label{sec: language exp analysis}
In task $\mathcal{K}{ \rightarrow }\mathcal{D}$, $\mathcal{BL}_2{ > }$\smallmethodname{} because of slight over-fitting on source domain.
Figure~\ref{fig:tsne all lang}(a-d) show the t-SNE plots of top four language tasks indicating classes being oriented on either half of the line like clustering.
\section{Conclusion} \label{sec:conclusions and future work}
In this paper, we have proposed a simple and direct approach that addresses the problem of unsupervised domain adaptation that is different from the standard distribution alignment approaches. In our approach, we jointly learn a Contradistinguisher (CTDR) on the source and target domain distribution in the same input feature space using contradistinguish loss for unsupervised target domain to identify contrastive features. We have shown that the contrastive learning overcomes the need and drawbacks of domain alignment, especially in tasks where domain shift is very high (e.g., language domains) and data augmentation techniques cannot be applied. Due to the inclusion of prior enforcing in the contradistinguish loss, the proposed unsupervised domain adaptation method \smallmethodname{} could incorporate any known target domain prior to overcome the drawbacks of skewness in the target domain, thereby resulting in a skew-robust model. We demonstrated the effectiveness of our model by achieving state-of-the-art results on all the visual domain adaptation tasks over eight different benchmark visual datasets and nine language domain adaptation tasks out of twelve along with the best mean test accuracy of all the twelve tasks on benchmark Amazon customer reviews sentiment analysis dataset. Specifically, the results in language domains reinforced the efficacy of \smallmethodname{} on being robust to high sparsity or high domain shift tasks that pose challenges to standard domain alignment approaches.

\section*{Acknowledgment}
The authors would like to thank Ministry of Human Resource Development (MHRD), Government of India, for their generous funding towards this work through UAY Project: {IISc 001} and {IISc 010}.



%
\balance

\bibliographystyle{IEEEtran}
\bibliography{IEEEabrv,cuda}




\end{document}